# Defense-guided Transferable Adversarial Attacks

Zifei Zhang , Kai Qiao, Jian Chen and Ningning Liang

**Abstract:** Though deep neural networks perform challenging tasks excellently, they are susceptible to adversarial examples, which mislead classifiers by applying human-imperceptible perturbations on clean inputs. Under the query-free black-box scenario, adversarial examples are hard to transfer to unknown models, and several methods have been proposed with the low transferability. To settle such issue, we design a max-min framework inspired by input transformations, which are benificial to both the adversarial attack and defense. Explicitly, we decrease loss values with inputs' affline transformations as a defense in the minimum procedure, and then increase loss values with the momentum iterative algorithm as an attack in the maximum procedure. To further promote transferability, we determine transformed values with the max-min theory. Extensive experiments on Imagenet demonstrate that our defense-guided transferable attacks achieve impressive increase on transferability. Experimentally, we show that our ASR of adversarial attack reaches to 58.38% on average, which outperforms the state-of-the-art method by 12.1% on the normally trained models and by 11.13% on the adversarially trained models. Additionally, we provide elucidative insights on the improvement of transferability, and our method is expected to be a benchmark for assessing the robustness of deep models.

**1. Introduction**

Recently, Deep Neural Networks (DNNs) play an important role in various tasks, including image classification [1,2], object detection [3,4], speech recognition [5], autonomous vehicles [6]. However, advanced neural networks are exceedingly vulnerability to intentionally crafted images and output wrong labels, by applying additive human-imperceptible perturbations, namely, adversarial examples. It is necessary to study the algorithms of generating adversarial examples for DNNs-based safety-critical applications like healthcare. In addition, learning adversarial examples provides a new idea to investigate the robustness of different DNNs and improve the robustness [7-10].

Adversarial attacks have drawn an increasing attention, and various methods have been proposed to find adversarial examples. Generally, according to if attackers have access to the model information, adversarial attacks can be categorized into two kinds, white-box and black-box attacks. Under the white-box setting, attackers can obtain the internal knowledge of the target model, e.g. network structure and parameters. In this setting, there are two types of attacks roughly, i.e., optimization-based [11,12] and gradient-based attacks [7, 13, 14, 15], and gradient-based attacks can compute gradient with one-step [13,14] and multi-step methods [7,15]. In the black-box manner, attackers have no access to target models directly, which meets real-world applications in practice. Considering whether attackers require to query target models, query-based and query-free attacks are proposed. By feeding images into target models and yielding probabilities or labels, attackers analyze probabilities or labels after querying, corresponding to score-based or boundary-based attacks, respectively. In essence, score-based attacks require large number of queries to obtain probabilities, which are employed to estimate gradients [16-18] and train surrogate models. Boundary-based attacks merely obtain hard-labels [19-21] with extremely limited information, thus need tremendous queries. They generate adversarial examples first, then shrink perturbations to approach decision boundary as soon as possible and remain adversarial simultaneously in hard way. Query-free attacks are more challenging but fit real-world systems. Such attacks mainly depend on an intriguing phenomenon of adversarial examples, namely, transferability, that is, adversarial examples generated by one model also keep adversarial property for other models. And we call such attacks as transfer-based attacks [22-24]. As shown in the experiment analysis [25], transferability derives from the similar decision boundary learned by various DNN models at a data point. From another viewpoint, the transferability helps to turn the black-box into white-box attack, and lessens the difficulty of attacks in the hard black-box manner. Recently, researchers also pay close attention to transfer-based black-box attack algorithms. The transferability is to adversarial examples what the generalizability is to models. By adopting such metaview, Dong et al. [22] extended momentum-based iterative attacks with momentum to boost adversarial examples' transferability (MIM), which is beneficial to escape from local maximum. Xie et al. [23] put forward Diversity Input Method (DIM) motivated by a data augmentation strategy, e.g. random resizing and padding. Oppositely, Xie et al. [26] ever presented a defense method to mitigate adversarial effects through feeding transformed inputs into the target model, by taking advantage of unrobust characteristic of transformed adversarial examples. They regarded models with the addition of transformations as a new model with two transformed layers, that is, resizing and padding. Transformation layers ahead of the source or target model contribute to adversarial attacks and defenses contemporaneously. In reality, diversity inputs attacks are arised from transformation-based defense. Analogously, as a classical defense method, adversarial training retrains models with clean inputs and adversarial examples, which is guided by adversarial attacks. Virtually, adversarial attacks and defenses help each other. Particularly, adversarial examples help to construct more robust models.

In this work, we focus on defense-guided transferable adversarial attacks, and consider how to further improve the transferability of adversarial examples. We propose a max-min framework, opposite to the min-max framework, and divide the process of generating adversarial examples into two parts, defense and attack. In the defense (minimum) procedure, new models are supposed to recognize adversarial examples with smaller loss values. Concretely, we select affline transformations like translation, scaling and rotation, and their ensembles. In the attack (maximum) procedure, we integrate our defense output into MIM at each iteration, which aims to maximize loss values as soon as possible. Taking another perspective, adversarial examples which are able to cheat multiple models are more likely to keep adversarial for other black-box models. Input transformations with random values increases model diversity and dynamically updates models, which can be regarded as model ensemble, too. Futhermore, ensemble transformations perform better than single transformation, due to considerable variety on models. Besides, we take advantage of the max-min framework to further confirm the value which we select leading to minimum loss at each iteration. Though our work is connected with DIM, yet they are diverse in several aspects: (1) we mainly concentrate on translation, scaling and rotation and their ensemble, while [23] focuses on the resizing and padding. (2) we integrate transformations with transverse ensemble whereas [23] incorporate these two transformations with the longitudinal ensemble. (3) we select transformed values with the max-min theory but [23] with random. In conclusion, we summary our contributions as follows.

1. We investigate transfer-based attacks and discover an interesting phenonmenon that defense methods are also benificial to such attacks, named defense-guided transferable adversarial attacks.
2. Inspired by the above-mentioned finding, we propose a max-min framework to provide a guide to generate more transferable adversarial examples effectively, which divides the procedure into two steps, decreasing loss with affline transformations as a defense in the minimum procedure, and then increasing loss with the momentum iterative algorithm as an attack in the maximum procedure.
3. Besides, we perform extensive experiments to improve the transferability of adversarial examples based on our max-min framework, which select transformed values corresponding to minimum loss values, to lower the loss as soon as possible in the defense procedure. And our study demonstrates a powerful attack even against adversarial training.

**2. Related Work**

In this section, we review related works on adversarial attack and defense methods.

*2.1. Attack Methods*

In the black-box setting, attackers cannot acquire information of target models directly. Transfer-based attacks can turn challenging black-box attacks into white white-box attacks based on the tranferability, where adversarial examples generated by a substitute model can fool an unkown model. In this study, we focus on gradient-based white-box attack methods to generate transferable adversarial examples.

**Fast Gradient Sign Method (FGSM).** FGSM [13] linearizes loss function and finds adversarial examples in the direction of increasing the loss with one-step update.

$$x^{adv} = x + \varepsilon \cdot \text{sign}\left(\nabla_x L\left(\theta; x, y^{true}\right)\right). \tag{1}$$

**Basic Iterative Method (BIM).** [15] extends FGSM to an iteration, which effectively improves the Attack Success Rate (ASR). The equation is

$$\begin{aligned} x_0^{adv} &= x, \\ x_{k+1}^{adv} &= \text{Clip}_x^\varepsilon \left\{ x_k^{adv} + \alpha \cdot \text{sign}\left(\nabla_x L\left(\theta; x_k^{adv}, y\right)\right) \right\}, \end{aligned} \tag{2}$$

Where $\alpha$ is the step size and meets $\alpha = \dfrac{\varepsilon}{K}$, and $K$ is the total iteration number, and $\text{Clip}_x^\varepsilon(\cdot)$ function requires adversarial examples within the $\varepsilon - \text{ball}$ of input images.

**Momentum Iterative Fast Gradient Sign Method (MIM).** MIM [22] integates the momentum into BIM, in order to escape from poor local maxima and improve transferability. And the updated procedure can be expressed as

$$\begin{aligned} g_{k+1} &= \mu \cdot g + \frac{\nabla_x L\left(\theta; x_k^{adv}, y^{true}\right)}{\left\|\nabla_x L\left(\theta; x_k^{adv}, y^{true}\right)\right\|_1}, \\ x_{k+1}^{adv} &= \text{Clip}_x^\varepsilon \left\{ x_k^{adv} + \alpha \cdot \text{sign}\left(g_{k+1}\right) \right\}, \end{aligned} \tag{3}$$

where $g_k$ is the accumulated gardient at the $k^{th}$ iteration, with the decay factor $\mu$.

**Diversity Inputs Method (DIM).** Xie et al. [23] propose a diverse input pattern to optimize the loss by the random resizing and padding transformations, which is motivated by data augmentation [27]. When combined with momentum, DIM generates more transferable adversarial examaples.

*2.2. Defense Methods*

Lots of defense methods have been raised against adversarial examples. There are two kinds of common defense methods, model modification [7,14,28,29] and inputs modification [26,30-32]. Among model modification methods, adversarial training is one of the most efficient defense methods experimentally [7], which augments training data with adversarial examples, and the model are formulared as the equation (4). Moreover, [14] proposed the ensemble adversarial training to improve the model robutness, with perturbations transferred from other models. Another category defense methods modificates inputs as a prepocessing technology, e.g. images denoising [32], images compression [31] and images transformation [26] by appending a transformed block before the target model. As a matter of fact, methods of inputs modification can be viewed as the model modification, which combine one layer with the old model as a new model.

$$\min_{\theta} \mathrm{E}_{\theta \sim \Psi} \left[ \max_{x^{adv}} L\left(\theta; x^{adv}, y^{true}\right) \right]. \tag{4}$$

## 3. Methods

*3.1. Motivation*

**Diversity Inputs Defense**. Constructing robust DNNs to adversarial examples is a challenging issue. Xie et. al [26] took low-level image transformations into consideration, i.e., additional resizing and padding two layers ahead of intrinsic models, which may be beneficial to destory adversarial examples' structrue. On the other hand, randomization in the inference procedure is hard for adversarial attacks. Thus, random resizing and padding are great transformations to enhance the capability against adversarial examples.

**Diversity Inputs Attack**. Although adversarial examples generated in the white-box setting can be misclassified by an unknown model as well, their transferability is limited. The reason is that adversarial examples are likely to overfit to the source model parameters and fall into the poor local maxima after several iterations. Conversely, the single-step attack [13] transfers better with the sacrifices of the large ASR in the white-box setting. It is difficult to alleviate the trade-off between the attack capacity of the white-box and black-box attacks. To solve this issue, Xie et. al [23] proposed diversity inputs with random resizing and padding as a data augmentation strategy.

**Diversity Inputs Defense and Attack**. As above mentioned, the input transformation like resizing and padding is not only valid for the defense, but also beneficial to the attack. In terms of diversity inputs defense method, [23] decreases loss values and largely drops ASR simultaneously. The reduction of loss is helpful for defense, by constrast, the growth of loss contributes to generating adversarial examples. In fact, diversity inputs attack profits from diversity inputs defense. As we can see, a class of such method is decribed as a max-min framework in formular (5), which is reverse to the min-max way of adversarial training.

$$\max_{x^{adv}} \mathrm{E}_{t \sim \Gamma} \left[ \min_{\theta} L\left(\theta; x^{adv}, y^{true}\right) \right], \tag{5}$$

where the distribution $\Gamma$ indicates the affine transformation sets including translation, scaling and rotation, which probably bring lower loss values through models. We call such methods as the defense-guided transferable adversarial attacks, based on our max-min framework.

*3.2. Affine Transformation*

Affine transformation is a linear mapping method that maintains both the collinearity relation between points and rates of distance along the line. In this paper, we apply a 2D affine transformation $\Gamma(x, y)$, and the point-wise transformation is as follow.

$$\begin{pmatrix} x' \\ y' \\ 1 \end{pmatrix} = \Gamma(x, y, a_{11}, a_{12}, a_{13}, a_{21}, a_{22}, a_{23}) = \begin{pmatrix} a_{11} & a_{12} & a_{13} \\ a_{21} & a_{22} & a_{23} \\ 0 & 0 & 1 \end{pmatrix} \begin{pmatrix} x \\ y \\ 1 \end{pmatrix}. \tag{6}$$

Instances of affine transformation involve translation、scaling、rotation and so on. There are six parameters in formular (6), which determine various transformations.

**Translation.** Translating images is a vector adding method that moves the original points using symmetric shifting vertically and horizontally. In fomular (7), $t_x$ determines the displacement along the $x$ and $t_y$ determines the displacement along the $y$. To simplify the transformation with fewer variables, we make the same $t_x$ and $t_y$.

$$\Gamma_t(x,y,t_x,t_y) = \begin{pmatrix} 1 & 0 & t_x \\ 0 & 1 & t_y \\ 0 & 0 & 1 \end{pmatrix} \begin{pmatrix} x \\ y \\ 1 \end{pmatrix}. \tag{7}$$

**Scaling.** Image zooming is a linear map which involves the geometric scaling up or down of image size. In fomular (8), $s$ specifies the scaling factor. Scaling is similar to resizing, but scaling is resizing the same size along the $x$ and $y$.

$$\Gamma_s(x,y,s) = \begin{pmatrix} s & 0 & 0 \\ 0 & s & 0 \\ 0 & 0 & 1 \end{pmatrix} \begin{pmatrix} x \\ y \\ 1 \end{pmatrix}. \tag{8}$$

**Rotation.** Rotation performs a geometric transformation, which rotates a user-specified angle around a point of origin. And in fomular (9), $\theta$ specifies the angle of rotation.

$$\Gamma_r(x,y,\theta) = \begin{pmatrix} \cos\theta & \sin\theta & 0 \\ -\sin\theta & \cos\theta & 0 \\ 0 & 0 & 1 \end{pmatrix} \begin{pmatrix} x \\ y \\ 1 \end{pmatrix}. \tag{9}$$

*3.3. Multi-transformation*

We explore two multi-transformations, which are stronger than the single transformation. They are average and minimum ensembles, and both of them are based on our proposed max-min framework. The brief introductions of them are demonstrated below.

**Average**. The mean loss ensemble method takes the average loss of several different transformations as its loss, named AIM. In fact, other than the single transformation, the method of average adds the model diversity further.

**Minimum**. Based on the average method, our minimum loss method is aimed at the thought of max-min mentioned above. Instead of random magnitudes in the average ensemble, it provides decided value each tranformation every iteration, corresponding to selecting the transformation magnitude with the minimum loss.

*3.4. Attack Algorithms*

In order to boost the transfer-based black-box attacks, we proposed defense-guided transferable adversarial attack methods, where our max-min framework works. And the pseudo-codes are summarized in **Algorithm 1**.

**4. Experiment**

In this section, we provide convictive experiment results to demonstrate validity of the proposed method. We first present experimental setup in Section 4.1. Then we explore the loss-reduction property of transformed adversarial examples in Section 4.2. Next, we report the attack results of single and multiple images transformations in Section 4.3 and 4.4, compared with baseline methods.

---

**Algorithm 1** Defense-Guided Transferable Adversarial Attacks Algorithm

---

**Input:** A sample pair of a clean example and its ground-truth label $(x, y)$; a classifier $f$ with its loss function $J$.

**Input:** The maximum magnitude of perturbation $\varepsilon$, the number of iterations $K$ and the average ensemble transformation $A_\Gamma(x, y, \theta, s, t)$.

**Output:** An adversarial example $x^{adv}$.

1: $\alpha = \varepsilon / K$ ;
2: $g_0 = 0$ ; $x_0^{adv} = x$ ; $\theta_0 = 0$ ; $s_0 = 1$ ; $t_0 = 0$ ;
3: **for** $k = 0$ **to** $K-1$ **do**
4:   Input $x_k^{adv}$ through the average ensemble tranformation $A_\Gamma(x, \theta_k, s_k, t_k)$ and obtain transformed $x_k^*$ ;
5:   Feed $x_k^*$ into $f$, and get the gradient of loss function $\nabla_x J(x_k^*, y^{true})$ ;
6:   Update $g_{k+1}$ by $g_{k+1} = \mu \cdot g_k + \dfrac{\nabla_x J(x_k^*, y^{true})}{\left\| \nabla_x J(x_k^*, y^{true}) \right\|_1}$ ;
7:   Refresh the new adversarial example $x_{k+1}^{adv} = x_k^* + \alpha \cdot \mathrm{sign}(g_{k+1})$ ;
8:   Renew the parameters of transformation as
$$\theta_{k+1}, s_{k+1}, t_{k+1} = \underset{\theta, s, t}{\mathrm{argmin}}\, J\left( A_\Gamma\left( x_{k+1}^{adv}, \theta_k, s_k, t_k \right), y^{true} \right) ;$$
9: **end for**
10: **return** $x^{adv} = x_{K-1}^{adv}$.

---

*4.1. Setup*

**Dataset**. We randomly select 1000 images belonging to 1000 categories from the ImageNet validation set [33], and all of them are correctly classified by our source models, since wrongly classified images are less meaningful to be studied on adversarial examples.

**Models**. We take six different models into consideration. They are three normally trained models, i.e., Inception-v3 [34], Inception-v4 and Inception-Resnet-v2 [35], and three are adversarially trained models, i.e., ens3-adv-Inception-v3, ens4-adv-Inception-v3 and ens-adv-Inception-Resnet-v2 [14]. In reality, aforesaid models with adversarial training strategy are hard to to attack. For convenience, we called them Inc-v3, Inc-v4, IncRes-v2, Inc-v3$_{ens3}$, Inc-v3$_{ens4}$ and IncRes-v2$_{ens}$, respectively.

**Attack Methods**. We adopt MIM and DIM as baseline attack methods, which are effective attacks with the high transferability. In our experiments, please note that DIM and our method combine with the momentum, since MIM is a useful and compatible tip.

**Implementation Details**. For the hyper-parameter settings, we set maximum perturbation as $\varepsilon = 16$ with pixel values in [0, 255], the number of iterations as $K = 10$ and the step size as $\alpha = 1.6$. For MIM, we take the default decay factor as $\mu = 1.0$. And as to DIM, we adopt the transformation probability $p = 0.5$. For detailed informations of our affine transformations, inputs can be randomly rotated angles $\theta$, with $\theta$ in [-18, 18], scaled ratio $s$, with $s$ in [0.9, 1.1], and translated step $t$, with $t$ in [-15, 15]. Our configuration settings are rational, since they largely influence adversarial examples, yet have a few effects on clean examples.

4.2. Loss-Reduction Property of Transformed Adversarial Examples

To validate the loss-reduction property of transformed adversarial examples, we generate 1000 adversarial examples with BIM, MIM and DIM attack methods, and transform them with translation, scaling and rotation, respectively. Then we feed them into the source model, Inc-v3, and then obtain the average loss over 1000 images.

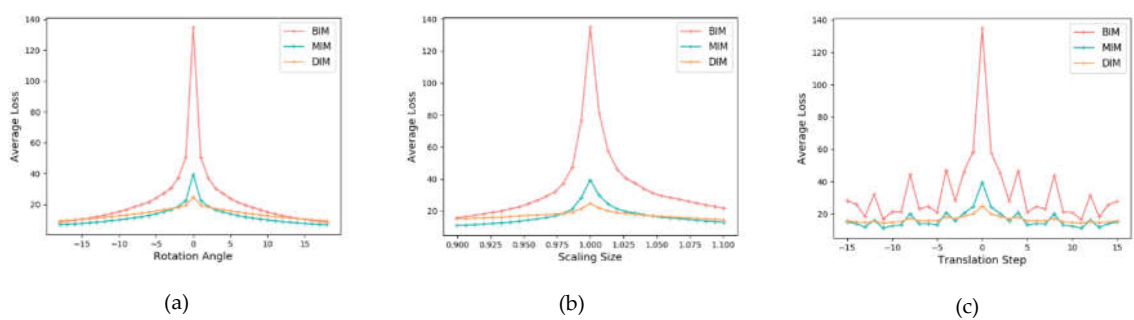

**Figure 1.** The curves of average loss with translation, scaling and rotation on Inc-v3.

As shown in Figure 1, when the rotation angle equals to 0, images don't rotate at all, as well as the scale size and translate step. And in such situation, the loss of DIM is the lowest, and that of MIM is also lower than BIM's, which illustrates the validity of the transformation. In addition, we can easily discover that non-transformed adversarial examples reach a maximum loss value, which confirms the defense effectiveness of diversity inputs. On the other hand, we observe that the curve of BIM drops a lot due to its adversarial examples falling into local maximum without optimization. However, the loss curve of DIM changes slowly, since DIM has experienced the input transformation before and it is insensitive to the following transformation. Besides, we find that the curves of scaling and translation are not symmetrical, which indicates the direction of such transformations also influence the loss. Another phenomenon is the unsmoothed translation curve, which may be that the loss value is impressible to the translation step.

*4. 3. Single-transformation Attack*

In this section, we study the ASR of transfer-based black-box attacks, and perform adversarial attacks with single transformations first. Adversarial examples are crafted by Inc-v3, Inc-v4 and IncRes-v2, which is useful to attack six models, including defensive adversarial traning models. Single image transformations with translation, scaling and rotation are abbreviated as TIM, SIM and RIM. In addition, we compare the transferability with baseline attacks, i.e., MIM and DIM.

We randomly choose an image from the ImageNet validation set and its ground-truth is 'castle'. Generally, attention maps of images indicate what the target models centralize. We utilize Grad-CAM++ to find the interesting regions related to labels. And the red the more, the region more important. In Figure 2, all adversarial examepls crafted by Inc-v3 are misclassified by IncRes-v2, and their labels are displayed at the bottom of attention maps. Different adversarial examples focus on different interests, and the central distances of salient regions also influence the labels. Adversarial examples crafted by MIM, DIM and SIM are predicted as 'Mosque', and their attention maps are close to that of the clean image. Obviously, salient regions of TIM and RIM move away, and they be classified as 'church' and 'theater curtain', respectively. On the other hand, the methods of DIM and SIM are similar, leading to the similar salient maps.

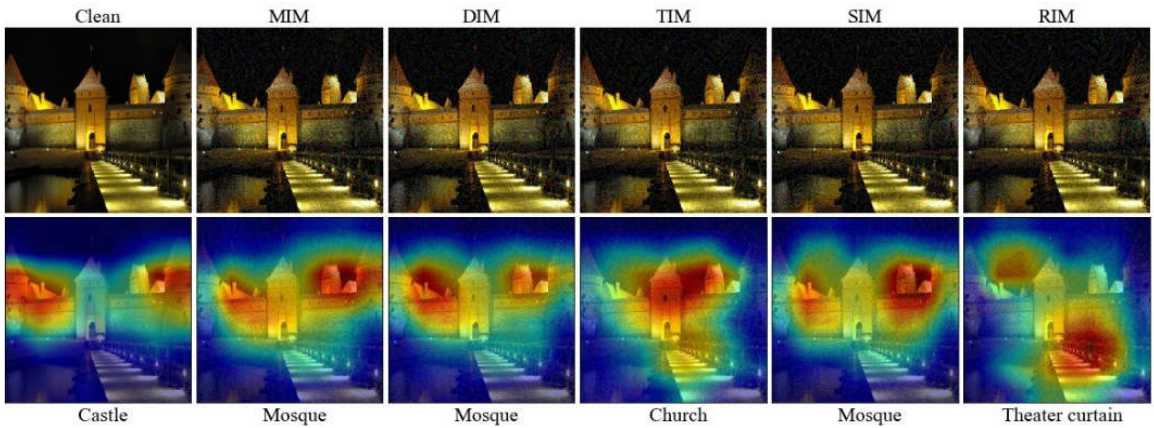

**Figure 2.** We visualize five adversarial examples and their attention maps. Row 1: demonstration of an original image and its adversarial examples crafted for Inc-v3 with five adversarial attacks, i.e., MIM, DIM, TIM, SIM and RIM. Row 2: We visualize the attention maps in Inception Resnet v2 with Grad-CAM++ [36], and display labels of images.

**Table 1.** The attack success rate (%) of adversarial attacks against six models, which are generated by MIM, DIM, TIM, SIM and RIM. The sign '*' indicates the white-box attack.

| Model | Attack | Inc-v3 | Inc-v4 | IncRes-v2 | Inc-v3$_{ens3}$ | Inc-v3$_{ens4}$ | IncRes-v2$_{ens}$ |
|---|---|---|---|---|---|---|---|
| Inc-v3 | MIM | **100**$^*$ | 51.9 | 51.4 | 17.5 | 14.5 | 9.1 |
| | DIM | 99.5$^*$ | 74.5 | 71.3 | 36.3 | 32.3 | 19.7 |
| | TIM | 98.8$^*$ | 75.6 | 70.7 | 26.0 | 25.0 | 13.0 |
| | SIM | 98.7$^*$ | 75.0 | 72.8 | 35.8 | 33.0 | 19.1 |
| | RIM | 98.3$^*$ | **79.8** | **76.0** | **37.1** | **33.7** | **20.8** |
| Inc-v4 | MIM | 60.14 | **99.6**$^*$ | 51.2 | 18.6 | 15.5 | 8.8 |
| | DIM | 77.2 | 99.0$^*$ | 70.3 | 31.3 | 28.2 | 16.4 |
| | TIM | 75.3 | 97.8$^*$ | 69.1 | 23.2 | 20.0 | 13.6 |
| | SIM | 77.3 | 97.5$^*$ | 71.3 | 30.8 | 27.6 | 16.6 |
| | RIM | **78.6** | 96.8$^*$ | **73.7** | **33.8** | **29.5** | **18.2** |
| IncRes-v2 | MIM | 63.1 | 55.1 | **98.2**$^*$ | 21.4 | 17.8 | 12.9 |
| | DIM | 73.5 | 72.4 | 95.6$^*$ | 39.3 | 33.4 | 26.8 |
| | TIM | 74.4 | 70.5 | 93.7$^*$ | 29.4 | 27.5 | 18.0 |
| | SIM | 73.7 | 72.4 | 94.5$^*$ | 38.2 | 34.3 | 25.4 |
| | RIM | **75.7** | **75.5** | 91.9$^*$ | **41.5** | **35.6** | **28.5** |

We observe the transferability of gradient-based attacks, and report the ASR of non-targeted adversarial examples in Table 1. Results in bold refer to the highest ASR for the current target-model. In the black-box setting, the attack dependent on the translation demonstrate transferability between MIM and DIM. The scaling transformation attack performs the nearly ASR to DIM, due to the similar transformation. Furthermore, our black-box attack based on the rotation performs best overall and outperforms DIM 3.35% on normally trained models and 1.67% on adversarially trained models. Although the results affirm the validity of our proposed methods, to some extent, the performance of adversarially trained models is supposed to be improved farther and we will focus on the stronger black-box attack method, like the multi-transformation attack.

*4.4. Multi-transformation Attack*

In this study, we discuss two types of multi-transformation attack. As described in Setion 3.3, the transformation with or without an explicit value corresponding to AIM-n or Max-min-n, respectively, where n represents the number of transformations. Although AIM-n with random values performs well, Max-min-n with specific values demonstrates better in transferability.

In Table 2, we report the ASR results of AIM-2 and AIM-3. For simplification, we regard attack methods with scaling, translation and rotation as S, T and R, respectively, and call scaling plus rotation as S+R. Intuitively, the method of AIM-2 is approximate 5.83% higher than single-transformation attacks in transferability. And AIM-2 with rotation plus scale (R+S) generates the most transferable adversarial examples among AIM-2. In addition, T+S+R consistently outperforms all AIM-2 attacks by a large margin under the black-box setting. For examples, when adversarial examples are crafted for Inc-v4, AIM-3 has ASR of 81.1% on IncRes-v2 and 23.4% on InRes-v2$_{ens}$, while strong AIM-2 with S+R only gets the corresponding ASR of 78.5% and 22.6%. AIM-3 improves ASR on normally training models effectively, which demonstrates the availability of max-min framework credibly, which outperforms the baseline attacks by 10.22% on the normally trained models. And compared with DIM, the black-box ASR of adversarially trained models boosts by 8.57%.

**Table 2.** The attack success rate (%) of adversarial attacks against six models, which are generated by S+T, R+T, R+S and T+S+R. '*' indicates the white-box attack.

| Model | Attack | Inc-v3 | Inc-v4 | IncRes-v2 | Inc-v3$_{ens3}$ | Inc-v3$_{ens4}$ | IncRes-v2$_{ens}$ |
| --- | --- | --- | --- | --- | --- | --- | --- |
| Inc-v3 | T+S | 99* | 81 | 78.8 | 36.6 | 34.6 | 18.2 |
| | T+R | 99.1* | 82.3 | 80.4 | 39.1 | 35.9 | 22.1 |
| | S+R | 98.9* | 82.6 | 80.8 | 43.3 | 40.3 | 25.1 |
| | T+S+R | **99.6*** | **86.1** | **83.5** | **44** | **41.9** | **26.6** |
| Inc-v4 | T+S | 81.2 | 98.5* | 76.8 | 31.6 | 29.0 | 18.1 |
| | T+R | 83.4 | 98.4* | 77.3 | 35.4 | 31 | 18.8 |
| | S+R | 83.6 | 98.5* | 78.5 | 37.1 | 32.6 | 22.6 |
| | T+S+R | **85.4** | **98.9*** | **81.1** | **40** | **34.6** | **23.4** |
| IncRes-v2 | T+S | 78.0 | 75.1 | **95.8*** | 40.5 | 36.5 | 26.9 |
| | T+R | 80.3 | 80.6 | 93.9* | 44.1 | 39.4 | 28.7 |
| | S+R | 81.4 | 80.8 | 93.8* | 49.5 | 42.9 | 32.3 |
| | T+S+R | **82.5** | **81.9** | 95.6* | **51.6** | **44.3** | **34.4** |

**Table 3.** The attack success rate (%) of adversarial attacks against six models, which are generated with Max-min-n based on methods in **Table 2**. '*' indicates the white-box attack.

| Model | Attack | Inc-v3 | Inc-v4 | IncRes-v2 | Inc-v3$_{ens3}$ | Inc-v3$_{ens4}$ | IncRes-v2$_{ens}$ |
| --- | --- | --- | --- | --- | --- | --- | --- |
| Inc-v3 | T+S+Max-min | 99.1* | 82 | 79.9 | 37.6 | 35.7 | 19.8 |
| | T+R+Max-min | 99* | 83.8 | 81.7 | 41.5 | 37.7 | 22.5 |
| | S+R+Max-min | 98.5* | 85.6 | 83.5 | 45.5 | 44.6 | 25.6 |
| | T+S+R+Max-min | **99.2*** | **88** | **85.9** | **46.9** | **43.4** | **27.5** |
| Inc-v4 | T+S+Max-min | 82.1 | 97.7* | 77.7 | 34 | 29.5 | 18.7 |
| | T+R+Max-min | 82.2 | 96.9* | 79.8 | 37.4 | 32 | 19.9 |
| | S+R+Max-min | 82.5 | 96.8* | 80.2 | 41.8 | 36.4 | 25 |
| | T+S+R+Max-min | **86.6** | **97.8*** | **82.6** | **43.8** | **38.5** | **27.5** |
| IncRes-v2 | T+S+Max-min | 79.9 | 78.3 | 92.8* | 41.7 | 38.1 | 28.3 |
| | T+R+Max-min | 81.7 | 81.6 | 92.1* | 46.1 | 41.2 | 29.8 |
| | S+R+Max-min | 82.9 | 81.9 | 91.6* | 51.0 | 44.9 | 33.7 |
| | T+S+R+Max-min | **85.1** | **83.6** | **92.9*** | **53.2** | **46.5** | **36.6** |

In this subsection, we extend our AIM-n to the Max-min-n, which generate strong adversarial examples against five black-box models. In Table 3, three types of Max-min-2 perform better than that of AIM-2, and among them, Max-min-2 with S+R transfers at the highest ASR, enstabling strong black-box attacks. It can be observed that Max-min-3 significantly outperforms all other attacks. For example, by attacking the IncRes-v2 model, the generated adversarial examples can fool challenging adversarially trained models like Inc-v3$_{ens3}$, Inc-v3$_{ens4}$ and IncRes-v2$_{ens}$ with the 53.2%, 46.5% and 36.6% ASR, respectively, and they are better than AIM-3 corresponding to the 51.6%, 44.3% and 34.4% ASR and DIM corresponding to the 39.3%, 33.4% and 26.8% ASR. And Max-min-3 consistently outstrips the baseline attack, DIM. Our Max-min-3 scheme is empirically validated to be suitable for tranfer-based adversarial attack in the black-box manner.

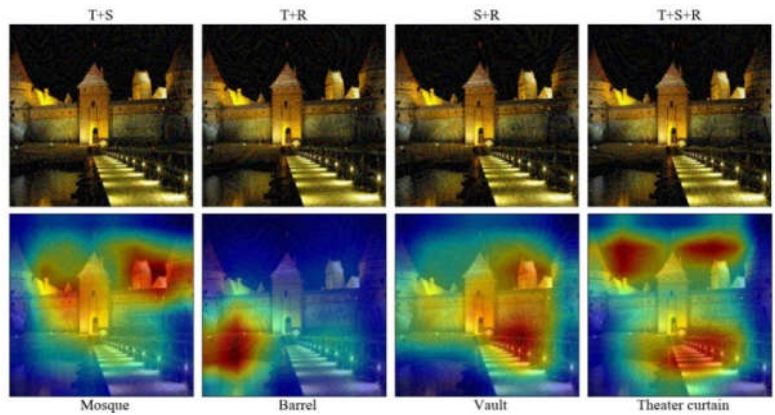

(a) AIM-n adversarial examples

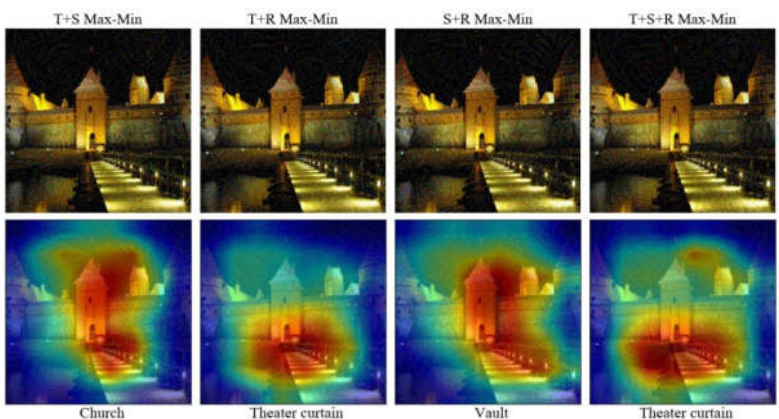

(b) Max-min-n adversarial examples

**Figure 3.** We visualize eight adversarial examples, with AIM-n and Max-min-n, and their attention maps. (a): demonstration of adversarial examples crafted for Inc-v3 with four average transformed attacks with AIM-n, i.e., T+S, T+R, S+R, T+S+R. Row 2 presents their attention maps of IncRes-v2 with Grad-CAM++, and display labels and probabilities of above images to show transferability of adversarial examples. (b): similar to (a), exhibition of adversarial examples crafted for Inc-v3 with four average transformed attacks with Max-min-n, i.e., T+S Max-min, T+R Max-min, S+R Max-min, T+S+R Max-min.

In Figure 3, adversarial examples, with AIM-n and Max-min-n, and their attention maps with Grad-Cam++ are demonstrated on (a) and (b), respectively. Intuitively, compared with the single transformation, attention maps of adversarial examples with AIM-n are more inclined to stay away from that of the original image, which are more likely to be transferable adversarial examples. From the view of concentricity, adversarial examples with our Max-min-n attack are characteristic, that is, attention maps of Max-min-n attacks are more centralized, which is beneficial to learn transferable adversarial examples.

## 5. Conclusions

In this paper, we propose a defense-guided transferable attack with our max-min framework, which conducts adversarial examples with a great transferability under the query-free black-box setting. Specifically, we are supposed to adopt the defense strategy to minimum loss values and attack methodology to maximum loss values orderly. In the defense procedure, affine transformations are taken ahead of initial models to build new models, which can be regarded as a method to enhance the model diversity as well. In the attack procedure, we apply iterative attacks with momentum to generate adversarial examples. In addition, we can utilize the max-min theory to restrain transformed values corresponding to minimum loss other than random, which also helps to facilitate transferability, e.g. DIM. The extensive experiments demonstrate the effectiveness of our proposed max-min framework and futher promote the transferability of adversarial examples. Experimentally, we show that our best black-box attack fools normally trained models at an 85.3% attack success rate and adversarially trained models at a 40.43% attack success rate on average, respectively. We expect that our max-min framework can guide to craft more transferable adversarail examples, enlighten more adversarial attack algorithms and be expected as a strong benchmark to assess the robustness models and defense mechanism.

**References**

1. Szegedy, C.; Liu, W.; Jia, Y.; Sermanet, P.; Reed, S.; Anguelov, D.; Erhan, D.; Vanhoucke, V.; Rabinovich, A. Going Deeper with Convolutions. In Proceedings of the 2015 IEEE Conference on Computer Vision and Pattern Recognition (CVPR 2015), Massachusetts, Boston, USA, 8-10 June 2015.
2. He, K.; Zhang, X.; Ren, S.; Sun, J. Identity Mappings in Deep Residual Networks. In Proceedings of the 14th European Conference on Computer Vision (ECCV 2016), Amsterdam, The Netherlands, 8-16 October 2016. pp. 630–645.
3. Zhang, Z.; Qiao, S.; Xie, C.; Shen, W.; Wang, B.; Yuille, A.L. Single-Shot Object Detection with Enriched Semantics. In Proceedings of the 2018 IEEE Conference on Computer Vision and Pattern Recognition (CVPR 2018), Salt Lake City, UT, USA, 19-21 June 2018. pp. 5813–5821.
4. Redmon, J.; Farhadi, A. YOLOv3: An Incremental Improvement. *arXiv* 2018, arXiv: 1804.02767.
5. Zheng, L.; Yang, Y.; Tian, Q. SIFT Meets CNN: A Decade Survey of Instance Retrieval. In *IEEE Transactions on Pattern Analysis and Machine Intelligence*, **2018**, 40, 1224–1244.
6. Bojarski, M.; Del Testa, D.; Dworakowski, D.; Firner, B.; Flepp, B.; Goyal, P.; Jackel, L.D.; Monfort, M.; Muller, U.; Zhang, J.; Zhang, X.; Zhao J. End to End Learning for Self-Driving Cars. *arXiv* 2016, arXiv: 1604.07316.
7. Madry, A.; Makelov, A.; Schmidt, L.; Tsipras, D.; Vladu, A. Towards Deep Learning Models Resistant to Adversarial Attacks. In Proceedings of the 6$^{th}$ International Conference on Learning Representations (ICLR 2018), Vancouver, Canada, 30–3 April-May 2018.
8. Ranjan, R.; Sankaranarayanan, S.; Castillo, C.D.; Chellappa, R. Improving Network Robustness against Adversarial Attacks with Compact Convolution. arXiv 2017, arXiv: 1712.00699.
9. Moosavidezfooli, S.; Fawzi, A.; Fawzi, O.; Frossard, P.; Soatto, S. Robustness of Classifiers to Universal Perturbations: A Geometric Perspective. In Proceedings of the 6$^{th}$ International Conference on Learning Representations (ICLR 2018), Vancouver, Canada, 30–3 April-May 2018.
10. Gao, J.; Wang, B.; Lin, Z.; Xu, W.; Qi, Y. DeepCloak: Masking Deep Neural Network Models for Robustness Against Adversarial Samples. In Proceedings of the 5$^{th}$ International Conference on Learning Representations (ICLR 2017 Workshop), Palais des Congrès Neptune, Toulon, France, 24-26 April 2017.
11. Szegedy, C.; Zaremba, W.; Sutskever, I.; Bruna, J.; Erhan, D.; Goodfellow, I.; Fergus. Intriguing properties of neural networks. In Proceedings of the 2$^{nd}$ International Conference on Learning Representations (ICLR 2014), Banff, AB, Canada, 14-16 April 2014.
12. 12.  Carlini, N.; Wagner, D. Towards Evaluating the Robustness of Neural Networks. In Proceedings of the 38$^{th}$ IEEE Symposium on Security and Privacy (S&P2017), San Jose, CA, US, 22-24 May 2017; pp. 39-57.
13. Goodfellow, I.J.; Shlens, J.; Szegedy, C. Explaining and Harnessing Adversarial Examples. In Proceedings of the 3$^{rd}$ International Conference on Learning Representations (ICLR 2015), San Diego, CA, US, 7-9 May 2015.
14. Tramer, F.; Kurakin, A.; Papernot, N.; Goodfellow, I.; Boneh, D.; Mcdaniel, P. Ensemble Adversarial Training: Attacks and Defenses. In Proceedings of the 6$^{th}$ International Conference on Learning Representations (ICLR 2018), Vancouver, Canada, 30–3 April-May 2018.
15. Kurakin, A.; Goodfellow, I.J.; Bengio, S. Adversarial Examples in the Physical World. In Proceedings of the 5$^{th}$ International Conference on Learning Representations (ICLR 2017 Workshop), Palais des Congrès Neptune, Toulon, France, 24-26 April 2017.
16. Chen, P.; Zhang, H.; Sharma, Y.; Yi, J.; Hsieh, C.J. ZOO: Zeroth Order Optimization Based Black-box Attacks to Deep Neural Networks without Training Substitute Models. In Proceedings of the 10th ACM Workshop on Artificial Intelligence and Security (AISec 2017), Dallas, TX, US, 3 Novermber 2017; pp 15-26.

17. Ilyas, A.; Engstrom, L.; Athalye, A.; Lin, J. Black-box Adversarial Attacks with Limited Queries and Information. In Proceedings of the 35th International Conference on Machine Learning, Stockholmsmassan, Stockholm, Sweden, 10-15 July 2018; pp. 2137-2146.
18. Tu, C.; Ting, P.; Chen, P.; Liu, S.; Zhang, H.; Yi, J.; Hsieh, C.; Cheng, S. AutoZOOM: Autoencoder-based Zeroth Order Optimization Method for Attacking Black--box Neural Networks. In Proceedings of the 33rd the Association for the Advance of Artificial Intelligence (AAAI 2019), Honolulu, Haiwii, US, 27-1 Jan-Feb 2019. pp. 742–749.
19. Brendel, W.; Rauber, J.; Bethge, M. Decision-Based Adversarial Attacks: Reliable Attacks Against Black-Box Machine Learning Models. In Proceedings of the 6th International Conference on Learning Representations (ICLR 2018), Vancouver, Canada, 30–3 April-May 2018.
20. Chen, J.; Jordan, M.I.; Wainwright, M.J. HopSkipJumpAttack: A Query-Efficient Decision-Based Attack. In Proceedings of the 41st IEEE Symposium on Security and Privacy (S&P2020), Virtual, 18-20 May 2020; pp.720-737.
21. Li, H.; Xu, X.; Zhang, X.; Yang, S.; Li, B. QEBA: Query-Efficient Boundary-Based Blackbox Attack. In Proceedings of the IEEE Conference on Computer Vision and Pattern Recognition (CVPR 2020), 16-18 June 2020; pp. 1221–1230.
22. Dong, Y.; Liao, F.; Pang, T.; Su, H.; Zhu, J.; Hu, X.; Li, J. Boosting adversarial attacks with momentum. In Proceedings of the IEEE Conference on Computer Vision and Pattern Recognition, Salt Lake City, UT, USA, 18–22 June 2018; pp. 9185–9193
23. Xie, C.; Zhang, Z.; Zhou, Y.; Bai, S.; Wang, J.; Ren, Z.; Yuille, A.L. Improving Transferability of Adversarial Examples with Input Diversity. In Proceedings of the IEEE Conference on Computer Vision and Pattern Recognition (CVPR 2019), Long Beach, CA, US, 16-20 June 2020; pp. 2730–2739.
24. Liu, Y.; Chen, X.; Liu, C.; Song, D. Delving into Transferable Adversarial Examples and Black-box Attacks. In Proceedings of the 5th International Conference on Learning Representations (ICLR 2017), Palais des Congrès Neptune, Toulon, France, 24-26 April 2017.
25. Tramer, F.; Papernot, N.; Goodfellow, I.; Boneh, D.; Mcdaniel, P.D. The Space of Transferable Adversarial Examples. ar*Xiv* 2017, arXiv: 1704.03453.
26. Xie, C.; Wang, J.; Zhang, Z.; Ren, Z.; Yuille, A.L. Mitigating Adversarial Effects Through Randomization. In Proceedings of the 6th International Conference on Learning Representations (ICLR 2018), Vancouver, Canada, 30–3 April-May 2018.
27. Shorten, C.; Khoshgoftaar, T. M. A survey on I`mage Data Augmentation for Deep Learning. In *Journal of Big Data*, **2019**, 6, 1-48.
28. Papernot, N.; McDaniel, P.; Wu, X.; Jha, S.; Swami, A. Distillation as a Defense to Adversarial Perturbations Against Deep Neural Networks. In Proceedings of the 37th IEEE Symposium on Security and Privacy (S&P2016), San Jose, CA, US, 23-25 May 2016; pp. 582–597.
29. Cisse, M.; Bojanowski, P.; Grave, E.; Dauphin, Y.; Usunier, N. Parseval Networks: Improving Robustness to Adversarial Examples. In Proceedings of the 5th International Conference on Learning Representations (ICLR 2017), Palais des Congrès Neptune, Toulon, France, 24-26 April 2017.
30. Shaham, U.; Garritano, J.; Yamada, Y.; Weinberger, E.; Cloninger, A.; Cheng, X.; Stanton, K.P.; Kluger, Y. Defending against Adversarial Images using Basis Functions Transformations. a*rXiv* 2018, arXiv:1803.10840.
31. Guo, C.; Rana, M.; Cisse, M.; Laurens, v.d.M. Countering Adversarial Images using Input Transformations. In Proceedings of the 6th International Conference on Learning Representations (ICLR 2018), Vancouver, Canada, 30–3 April-May 2018.
32. Xie, C.; Wu, Y.; Maaten, L.v.d.; Yuille, A.L.; He, K. Feature Denoising for Improving Adversarial Robustness. In Proceedings of the IEEE Conference on Computer Vision and Pattern Recognition (CVPR 2019), Long Beach, CA, US, 16-20 June 2020; pp. 2730–2739; pp. 501-509.
33. Deng, J.; Dong, W.; Socher, R.; Li, L.J.; Li, F.F. ImageNet: A large-scale hierarchical image database. In Proceedings of the 2009 IEEE Conference on Computer Vision and Pattern Recognition (CVPR 2009), Miami Beach, Florida, US, 20-25 June 2009; pp. 248–255.
34. Szegedy, C.; Vanhoucke, V.; Ioffe, S.; Shlens, J.; Wojna, Z. Rethinking the Inception Architecture for Computer Vision. In Proceedings of the IEEE Conference on Computer Vision and Pattern Recognition (CVPR 2016), Las Vegas, Nevada, US, 26-1 June-July 2016; pp. 2818–2826.

35. Szegedy, C.; Ioffe, S.; Vanhoucke, V.; Alemi, A. Inception-v4, Inception-ResNet and the Impact of Residual Connections on Learning. In Proceedings of the 30th the Association for the Advance of Artificial Intelligence (AAAI 2016), Phoenix, Arizona, US, 12-17 Feb 2016; pp.4278-4284
36. Chattopadhyay, A.; Sarkar, A.; Howlader, P.; Balasubramanian, V.N. Grad-CAM++: Improved Visual Explanations for Deep Convolutional Networks. In Proceedings of the 2018 IEEE Winter Conference on Applications of Computer Vision (WACV 2018), Lake Tahoe, CA, US, 12-15 March 2018; pp. 839–847.